\documentclass[conference]{IEEEtran}
\usepackage{helvet}
\usepackage{titlesec}
\usepackage[dvipsnames, table]{xcolor}
\usepackage[]{algorithm2e}
\usepackage{algpseudocode}
\usepackage{listings}
\usepackage{subcaption}
\usepackage{amsmath}
\usepackage{cite}
\usepackage{graphicx}
\usepackage{enumitem}
\usepackage{booktabs} 
\usepackage{multirow} 
\ifCLASSINFOpdf
\else
\fi

\begin{document}
%


\title{Aesthetic Preference Prediction in Interior Design: Fuzzy Approach}














\author{\IEEEauthorblockN{Ayana Adilova \IEEEauthorrefmark{1},
Pakizar Shamoi\IEEEauthorrefmark{2}}
\IEEEauthorblockA{School of Information Technology and Engineering \\
Kazakh-British Technical University\\
Almaty, Kazakhstan\\
Email: \IEEEauthorrefmark{1}ay\_adilova@kbtu.kz,
\IEEEauthorrefmark{2}p.shamoi@kbtu.kz,
}
}


%


\maketitle



%
\IEEEpeerreviewmaketitle

\begin{abstract}

Interior design is all about creating spaces that look and feel good. However, the subjective nature of aesthetic preferences presents a significant challenge in defining and quantifying what makes an interior design visually appealing. The current paper addresses this gap by introducing a novel methodology for quantifying and predicting aesthetic preferences in interior design. Our study combines fuzzy logic with image processing techniques. We collected a dataset of interior design images from social media platforms, focusing on essential visual attributes such as color harmony, lightness, and complexity. We integrate these features using weighted average to compute a general aesthetic score. Our approach considers individual color preferences in calculating the overall aesthetic preference. We initially gather user ratings for primary colors like red, brown, and others to understand their preferences. Then, we use the pixel count of the top five dominant colors in the image to get the color scheme preference. The color scheme preference and the aesthetic score are then passed as inputs to the fuzzy inference system to calculate an overall preference score. This score represents a comprehensive measure of the user's preference for a particular interior design, considering their color choices and general aesthetic appeal. We used the 2AFC (Two-Alternative Forced Choice) method to validate our methodology, achieving a notable hit rate of 0.7. This study can help designers and professionals better understand and meet people's interior design preferences, especially in a world that relies heavily on digital media. 





\end{abstract}


\section{Introduction}

Interior design is a complex and subjective field that requires a deep understanding of aesthetics and clients' preferences. Nowadays, many people struggle to find a flexible and attractive design. For example, they struggle to choose the right color or placement of furniture. 

While designers have access to many tools for visualizing interior spaces, there are not enough computational methods to objectively analyze the aesthetic qualities of interior design. Understanding what people find beautiful in interior design can be challenging, especially since everyone has their own tastes, and aesthetic preferences are subjective in nature \cite{Ren_2017_ICCV}. 

Aesthetic evaluation of interior design is studied and based on visual features \cite{disc1}. Aesthetic visual quality and aesthetic quality classification of photographs are studied and analyzed by color harmony \cite{new15}, \cite{new24}, \cite{new25}. Based on this, we can conclude that color harmony is indeed a crucial factor in determining the perceived quality of a photo. It plays a significant role in creating a visually pleasing composition and can greatly impact the overall aesthetic appeal of an image \cite{new_harm}. However, there is a lack of visual features from interior design images and some challenges in computational image aesthetics \cite{new}.

The current research presents a novel methodology for estimating interior design aesthetic preferences to eliminate this gap. In our work, fuzzy logic and image processing methods are combined.  Fuzzy logic allows for more flexible reasoning than traditional binary logic. It is suitable for dealing with subjective and imprecise data like individual preferences and aesthetics. We collect a dataset of popular interior designs from Instagram, which ensures that the designs are relevant and of high interest to the public. Our method calculates the overall aesthetic preference while considering individual color choices. We employ a fuzzy inference system to deduce users' personalized preferences for specific interior designs, considering users' color preferences and general aesthetic appeal. The general aesthetics score is found by integrating visual elements like color harmony, lightness, and complexity. Color preferences are, in turn, obtained by collecting user evaluations for primary colors and counting the pixels of the top five prominent colors in the image.




The contributions of our study are as follows:
 \begin{itemize}
 \item Application of Fuzzy Logic to interior design aesthetics. We propose a novel way to quantify subjective elements like color harmony, complexity, and brightness and integrate them with user preferences using fuzzy inference.
\item Account for individual differences in aesthetic preference estimation. The fuzzy inference system takes users' color preferences as an additional input.
\item We gathered our own collection of the latest and most liked interior designs from social media. This is important due to the dynamic nature of design trends. Our analysis specifically focuses on the designs that have gained prominence in the years 2023 and 2024.


 \end{itemize}




The paper is structured as follows. Section I is
this introduction. An overview of literature related to computational aesthetics is presented in Section II. Methods, including data collection, aesthetics evaluation, and fuzzy techniques are covered in Section III. Section IV presents the experimental results. Concluding remarks are provided in Section V, along with recommendations for future development.


\section{Related Work}
The current section provides an overview of computational aesthetics and aesthetic preference research. 

Several computational methods have been developed to analyze and evaluate visual media. Birkhoff applied the theory of aesthetic measure \cite{Wang2016}, \cite{Panetta2019}, and the other study applied the theoretical framework for understanding aesthetic experiences \cite{new18}. The paper measures the identity, similarity, contrast, and ambiguity of pictures of websites and graphic design and marketing. They developed a method of aesthetic measure \cite{new26}, \cite{new27}. 

The measure of aesthetic preference \cite{aesthpref} is evaluated and explored. Peel's study \cite{Peel1946} used a system where people ranked pictures and designs by how much they liked them. The use of different ways to measure the aesthetic preferences of users encourages more studies on this subject \cite{Palmer2013}. A study \cite{8970458} introduced a new deep learning framework that incorporates personality traits to assess image aesthetics both generally and personally. \cite{Ren_2017_ICCV} propose a novel approach for learning a personalized image aesthetics model with a residual-based model adaptation, which outperforms existing methods. Existing methods in Personalized Image Aesthetics Assessment (PIAA) rely on centralized machine learning techniques, which can potentially compromise the privacy of sensitive image and rating information. To improve PIAA, \cite{10219861} introduce an innovative approach: the Federated Learning-empowered Personalized Image Aesthetics Assessment (FedPIAA). This method employs a model design that effectively discerns patterns in image aesthetics and individual user preferences, all within a privacy-conscious framework.

Some studies have demonstrated that there is a linear relationship between complexity and aesthetic preference \cite{DELPLANQUE2019146}, and others stated that aesthetic preference can be predicted from a mixture of low- and high-level visual features\cite{Iigaya2021}. 

Computational aesthetics analysis has been used to analyze and compare the aesthetics of different art forms, such as paintings, photographs, and digital art and implement personalized image aesthetic assessment \cite{Zhou2021}, \cite{Brielmann2021}, \cite{new13}, \cite{new14}. Design aesthetics was evaluated by a phenomenological approach \cite{new0}, \cite{new16} and by the naturalness and aesthetic appeal of color \cite{new2}. Other works use CNN \cite{new1}, \cite{new12}, or machine learning algorithms to find aesthetic perception and better understand why people like or dislike visual media \cite{new8}. Some studies have employed a semantic approach \cite{new6}, \cite{new5}, \cite{new9}, reviews, experiment results \cite{new4}, \cite{new5} and models developed \cite{new10}, \cite{new17}.

Similarly, deep learning techniques were used to rate the image aesthetics \cite{new23} and design-awards datasets implemented to build in computational aesthetic evaluation model \cite{new19}. In their study, \cite{new7} investigated the use of color combinations to enhance the aesthetic quality of images, while \cite{Shamoi2022} focused on image retrieval techniques.

Other studies quantify aesthetics using color harmony \cite{Wu2011}, \cite{Shamoi2016}. Similarly, algorithms have been developed for detecting color harmony by analyzing the color distribution and contrast in images \cite{Gao2020}, \cite{Schloss2011}, \cite{new20}. Using SVM, a model was created using the local color harmony features and the overall color harmony. Other techniques include analyzing visual media's texture, shape, and composition. 

Next, \cite{new22} applied quantifying beauty system and aesthetic measurement approach \cite{new21}. The study conducted by \cite{new3} aimed to analyze the differences in color preferences across different cultures. Another researcher created a Python library for analyzing the aesthetics of visual media in social science research, as described in \cite{new_python}.

Several studies consider individual differences in perception of aesthetics \cite{DELPLANQUE2019146}. Some studies are working with datasets of images, while others are conducting surveys and developing design methodologies \cite{123568}, \cite{new26}, \cite{new27}. 
 
Given that the concept of "visual appeal" is inaccurate and difficult to pinpoint, automatic evaluation of the aesthetics of visual media is a difficult task. The task becomes even more challenging in the case of personalized aesthetic assessments.


\section{Methods}
\subsection{Data Collection and Description}
\begin{table}[tb]
\centering
\caption{Dataset}
\label{datasett}
\begin{tabular}{cccccc}
\toprule
\textbf{Image ID} & \textbf{Likes} & \textbf{Color Harmony} & \textbf{Lightness} & \textbf{Complexity} \\
\midrule
1   & 807  & 97.41 & 5 & 149 \\
2   & 2558 & 99.63 & 6 & 78  \\
3   & 154  & 96.75 & 6 & 149 \\
4   & 140  & 100   & 5 & 140 \\
5   & 604  & 98.98 & 4 & 79  \\
6   & 907  & 99    & 6 & 332 \\
7   & 308  & 100   & 5 & 121 \\
8   & 502  & 98.98 & 6 & 228 \\
9   & 1533 & 97.86 & 5 & 137 \\
10  & 443  & 98.63 & 6 & 346 \\
........& .........&........& .......& .......\\
100 & 384  & 100   & 5 & 296 \\
\bottomrule
\end{tabular}
\end{table}

We gathered a hundred images from a public Instagram page related to interior design and recorded the number of likes each image received. Social media likes generally reflect user interest or engagement with the content. We hypothesized that users would favor an image with more likes than an image with fewer likes.

We standardized our interior design image dataset by resizing all images to a uniform dimension of 200x200 pixels and ensuring they were in RGB mode. Then, we extracted features of each image - Color Harmony, Complexity, and Lightness (discussed in the following subsections). Various techniques have been developed to analyze color harmony, image complexity, and brightness by examining the variance in grey-level intensities. 

\begin{figure*}[htbp]
\centerline{\includegraphics[width=\textwidth]{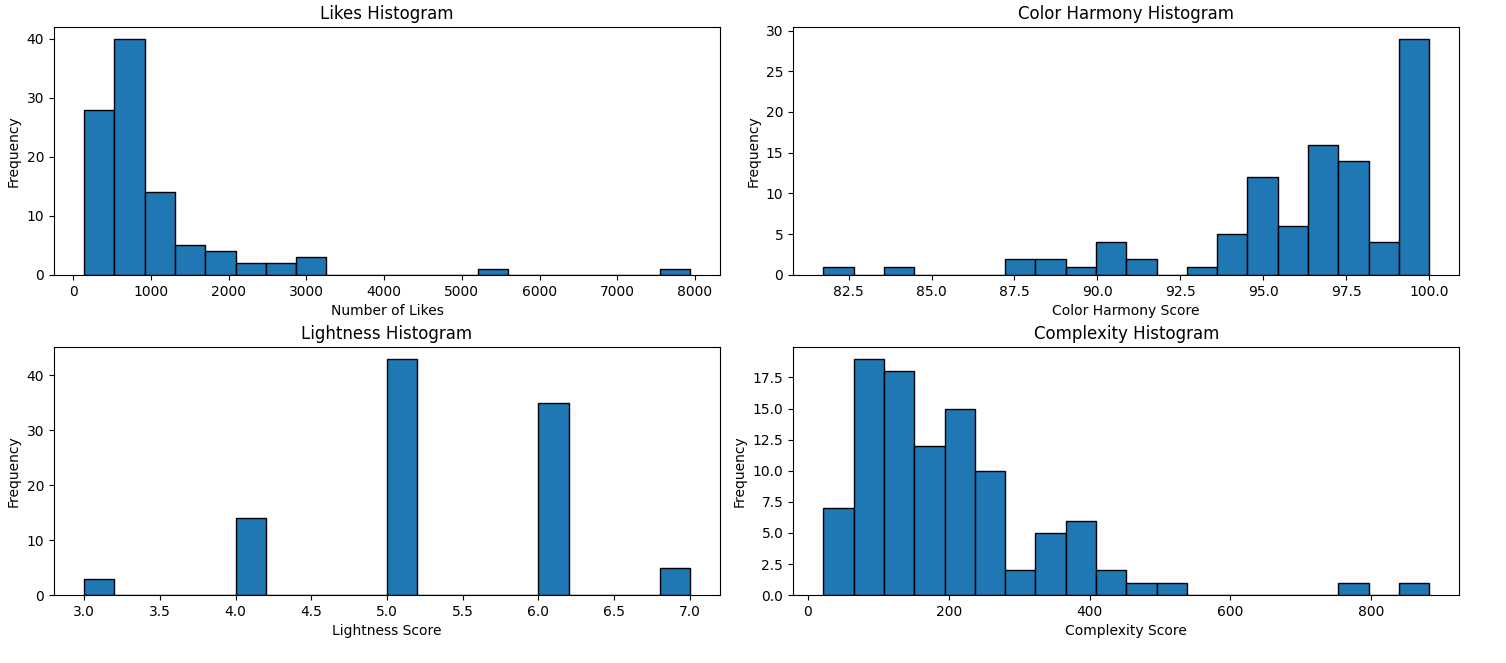}}
\caption{Histograms of visual features in the dataset}
\label{fig_hist}
\end{figure*}

So, besides the fields such as \textit{ImageID} and \textit{Likes}, the dataset includes scores for \textit{Color Harmony}, \textit{Lightness}, and \textit{Complexity} of each image. The dataset is presented in Table \ref{datasett}. The histograms for each feature are presented in Fig. \ref{fig_hist}.


\subsection{Fuzzy Sets and Logic Theory}
This section overviews the fundamental terms from the fuzzy theory we use in our research \cite{fss}.

Fuzzy logic provides a formal framework for dealing with uncertainty and imprecision in decision-making processes. It represents and manipulates ambiguous concepts and enables reasoning based on approximate or incomplete information. Lotfi A. Zadeh is considered the founder of fuzzy logic and has made significant contributions to the field \cite{Zadeh1965}, \cite{Zadeh1975}.

\begin{figure*}[tb]
  \begin{subfigure}{0.33\textwidth}
    \includegraphics[width=\linewidth]{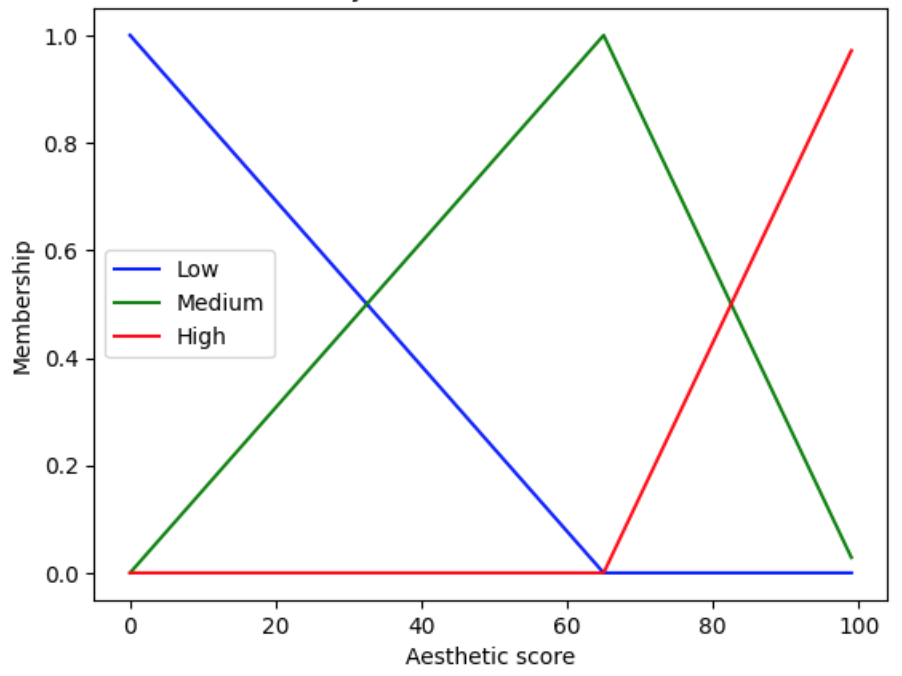}
  \end{subfigure}%
  \begin{subfigure}{0.33\textwidth}
    \includegraphics[width=\linewidth]{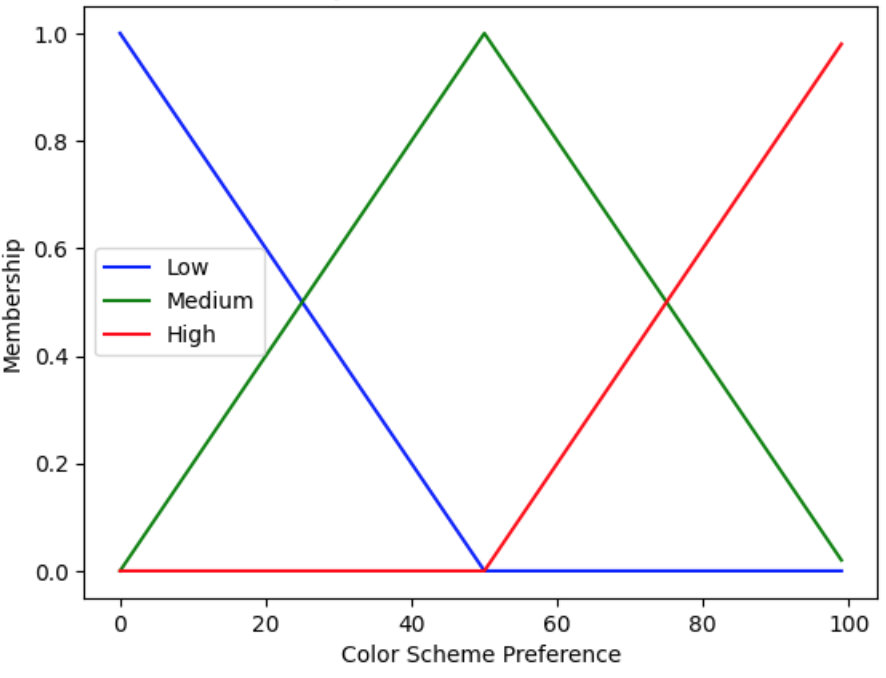}
  \end{subfigure}%
    \begin{subfigure}{0.33\textwidth}
    \includegraphics[width=\linewidth]{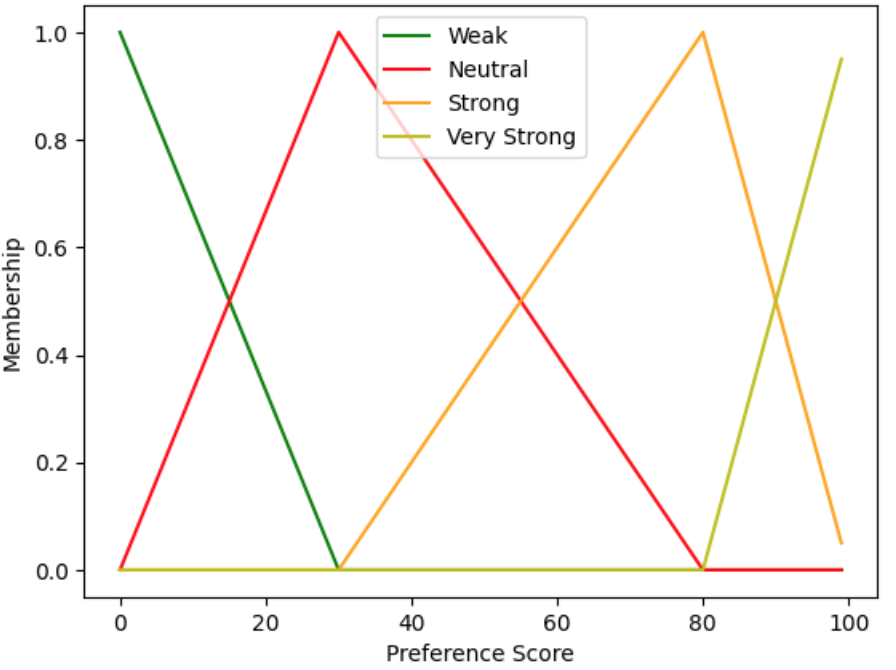}
  \end{subfigure}
 \caption{Input fuzzy sets for \textit{Aesthetic Score}, \textit{Color Scheme Preference} and Output fuzzy sets for \textit{Preference }} 
\label{many_examples_palettes}
\end{figure*}

\subsubsection{Membership Functions and Fuzzy Sets}
In a formal context, a membership function for a fuzzy set defined on the universe of discourse \( X \) can be expressed as \( \mu_A: X \rightarrow [0, 1] \). This function assigns each element of \( X \) a value within the range of 0 to 1, denoted as the membership value or degree of membership. This value measures how strongly an element in \( X \) belongs to the fuzzy set \( A \). In this context, \( X \) represents the universal set, while \( A \) represents the fuzzy set derived from \( X \).

Imagine a color gradient where it's hard to tell where one color ends and another begins – that's what a fuzzy set is like, and it's ideal for modeling aesthetic preferences that don't have hard boundaries \cite{fss}. It's a way of quantifying preferences that feels more human-consistent.

\subsubsection{Linguistic Variables}

Linguistic variables are those whose values are words or sentences in a natural or artificial language rather than numbers \cite{Zadeh1975} (languages that emerge in computer simulations, robot interactions, etc.). For instance, the term \textit{Color Harmony} can be used to indicate a linguistic variable with possible values such as \textit{Neutral, Weak, Strong,} and so on.

In our case, we have three linguistic variables - \textit{Aesthetic Score}, \textit{Color Scheme Preference}, and Output fuzzy set \textit{Preference }. For the sake of simplicity, we use triangular membership functions as shown in Fig. \ref{many_examples_palettes}. A fuzzy partition was done after analyzing the dataset's aesthetic scores distribution.

\subsubsection{Fuzzy Operations}
The $\alpha $-cut (Alpha cut) is a crisp set that includes all the members  of the given fuzzy subset f whose values are not less than $\alpha $ for  
$\alpha  \quad  \le $ 1:
\[
f_{\alpha}  = \{x:\mu _{f} (x) \ge \alpha \}
\]
To connect $\alpha $-cuts and 
set operations (A and B are fuzzy sets):
\[
(A \cup B)_{\alpha}  = A_{\alpha}  \cup B_{\alpha}  ,
\quad
(A \cap B)_{\alpha}  = A_{\alpha}  \cap B_{\alpha}  
\]
\subsubsection{Fuzzy Rules}
Fuzzy rules are given as a  collection of sets of rules to decide on the classification of an input or determining an output. Fuzzy rules are given as in the following:

\textit{if  (\textit{input1 is membership function (MF)1}) \\
\textit{and/or } (\textit{input2 is membership function (MF)}) \\
\textit{and/or } (\textit{input3 is membership function (MF)}) \\
\textit{then } (\textit{output membership function (MF)})}

For example, if \textit{color harmony }is \textbf{low}, and \textit{lightness} is \textbf{low}, and \textit{complexity} is \textbf{high}, then\textit{ aesthetic appeal} is \textbf{weak}.

Note that \textit{low, high, weak} are fuzzy sets here, while \textit{lightness}, \textit{complexity}, \textit{color harmony}, and \textit{aesthetic appeal} are linguistic variables.
 
\subsubsection{Fuzzy Inference}

Fuzzy inference is a method used in fuzzy logic systems to interpret input data and make decisions based on a set of rules \cite{fismamdani}. It is useful when decisions need to be made with imprecise information. The process includes four main steps:
\begin{enumerate}
    \item \textbf{Fuzzification}. Converting actual data inputs into fuzzy sets.
    \item \textbf{Rule Evaluation}. Applying 'IF-THEN' rules to these sets.
    \item \textbf{Aggregation}. Combining the outputs from all rules in the knowledge base.
    \item \textbf{Defuzzification}. Converting the fuzzy output back into a number.
\end{enumerate}

\subsection{Proposed approach}

\begin{figure*}
  \includegraphics[width=\textwidth]{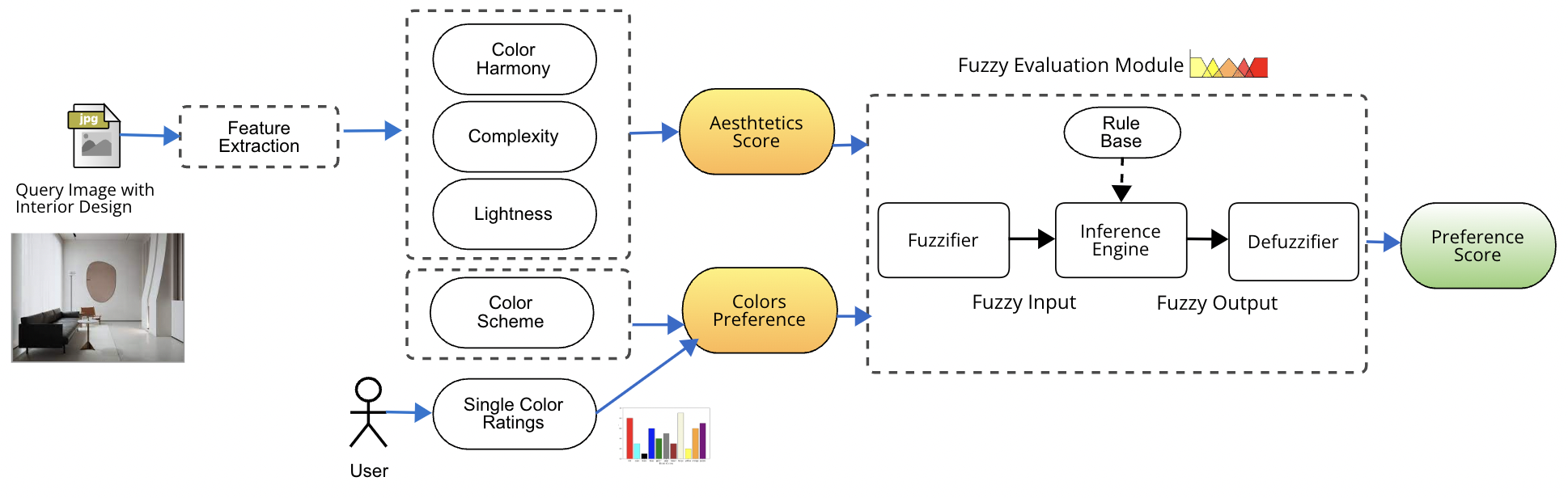}
\caption{High-level representation of the method. It has two logical parts - Feature Extraction and Fuzzy Evaluation Modules.} 
\label{figFuzzy}
\end{figure*} 

Our methodology integrates color harmony, brightness, and complexity in evaluating general aesthetic levels. These are the specific elements of interior design being considered. \textit{Color Harmony (CH)} refers to how well different colors in an image work together, \textit{Lightness (L)} refers to how light the image is, and \textit{Complexity (C)} is related to the visual overload of the design elements within the space. 

The proposed aesthetic preference evaluation method is presented in Fig. \ref{figFuzzy}:
\begin{itemize}
  \item Performing Feature Extraction
    \item Determining Aesthetic Score using obtained features
    \item Evaluating Color Preferences of User
    \item Evaluating aesthetic preferences through fuzzy logic rules, using the Aesthetic Score and Color Preference as inputs.
\end{itemize}

\subsubsection{Feature Extraction}
\paragraph{Color Harmony}
We utilized the FHSI color model library \cite{fss} to evaluate the color harmony level. We transformed the images into the HSI model (Hue, Saturation, Intensity) followed by the conversion into the FHSI model (fuzzy color histogram). We obtain a CH level ranging from 0 to 100 for each image in the dataset.

\begin{figure}[tb]
\centerline{\includegraphics{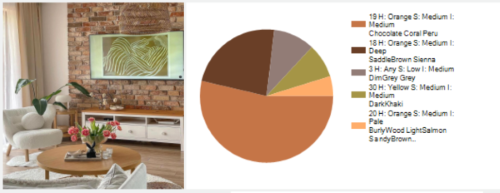}}
\centerline{\includegraphics{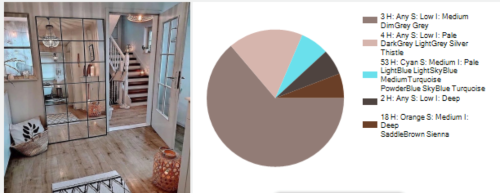}}
\caption{Harmony evaluation using the dominant fuzzy color scheme of interior designs. Method adapted from \cite{fss}. Top image (ID=16) has a \textit{Harm = 94.34}, bottom image (ID=18) has \textit{Harm = 98.15}}
\label{figharm}
\end{figure}

The harmony evaluation in FHSI compares the given image's fuzzy color scheme with statistically harmonious fuzzy color-based pallets extracted by grouping images with similar color compositions \cite{fss}. The more similar the image color scheme to a harmonious palette, the higher its harmony level. Fig. \ref{figharm} illustrates an example of the harmony evaluation of interior designs.

\begin{figure}
\centerline{\includegraphics{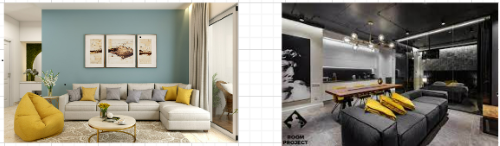}}
\caption{Lightness Estimation. The left image has a calculated lightness value of 6, and the right image's lightness is 3}
\label{Lightness}
\end{figure}

\paragraph{Lightness}
In estimating lightness, we use a generalized algorithm to detect the brightness of any image \cite{snscrape}. This algorithm takes an image as input and gives a score between (0-10) as output (zero being low bright and ten being high bright). It utilizes the PIL library (Python Imaging Library) to open and analyze the image:
\begin{itemize}
      \item Create bins of 10 levels from 0 to 255, for quantization of results between 10 levels;
      \item Get the average pixel level for each channel;
    \item Apply formula of perceived brightness, i.e. 
    $brightness = \sqrt{0.299 R^2 + 0.587 G^2 + 0.114 B^2}$
    \item Obtain the level of brightness between (1 to 10).
The example can be seen in Fig. \ref{Lightness}.
\end{itemize}

\paragraph{Complexity}
\begin{figure}
\includegraphics[width=0.5\textwidth]{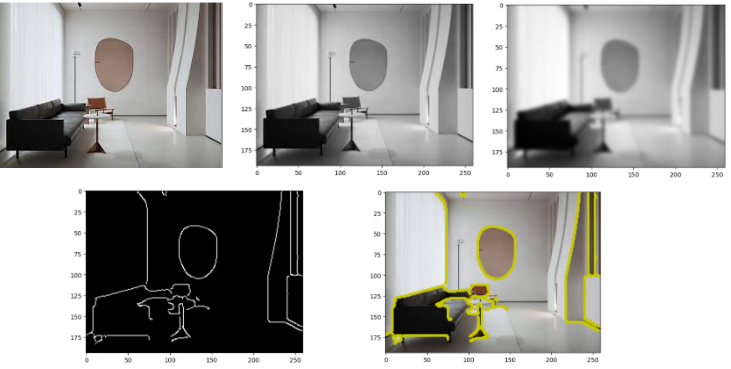}
\caption{{Complexity Estimation}}
\label{COMPLEXITY}
\end{figure}

Visual complexity is about the number of objects or elements in the interior space. The more objects there are, the higher the visual complexity and possibly the perceived visual weight. When an interior space has more objects, more visual information must be processed. This increased visual complexity can affect the design's overall balance and aesthetics. 

Fig.\ref{COMPLEXITY} provides an example of visual complexity obtained from an image. The image is first read and made grayscale. Then, to reduce noise, it is blurred. The Canny algorithm is then used for edge detection. The number of items and the contours are then calculated.

\begin{figure}
  \includegraphics[width=0.48\textwidth]{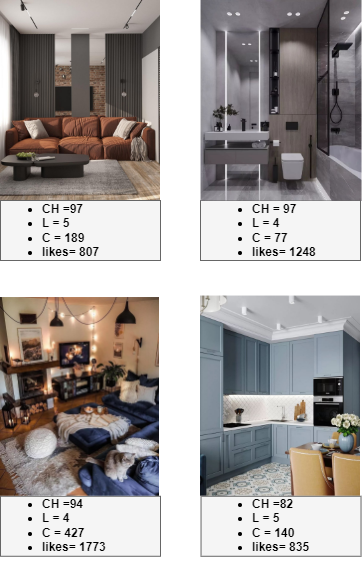}
\caption{Examples of images and calculated values - \textit{CH, C, L}.}
\label{calcValues}
\end{figure}

Finally, we can calculate \textit{CH, C, L }for every image. The examples of images and the calculated values are presented in Fig.\ref{calcValues}.

\subsubsection{Aesthetic Score}
\begin{table*}[tb]
\centering
\caption{Results Aesthetic score, Normalized CH, L, C, and Simplicity}
\label{dataset_aes}
\begin{tabular}{rrcccccccc}
\toprule
\textbf{Image ID} & \textbf{Likes} & \textbf{Color Harmony} & \textbf{Lightness} & \textbf{Complexity} & \textbf{CH\_norm} & \textbf{L\_norm} & \textbf{C\_norm} & \textbf{Simplicity\_norm} & \textbf{Aesthetics Score} \\
\midrule
1 & 807  & 97.41 & 5 & 149 & 0.86 & 0.50  & 0.15 & 0.85 & 0.68 \\
2 & 2558 & 99.63 & 6 & 78  & 0.98 & 0.75 & 0.07 & 0.93 & 0.85 \\
3 & 154  & 96.75 & 6 & 149 & 0.82 & 0.75 & 0.15 & 0.85 & 0.79 \\
4 & 140  & 100   & 5 & 140 & 1    & 0.50  & 0.14 & 0.86 & 0.72  \\
5 & 604  & 98.98 & 4 & 79  & 0.94 & 0.25 & 0.07 & 0.93 & 0.59 \\
6 & 907  & 100   & 6 & 332 & 1    & 0.75 & 0.36 & 0.64 & 0.79  \\
7 & 308  & 100   & 5 & 121 & 1    & 0.50  & 0.12 & 0.88 & 0.72   \\
8 & 502  & 98.98 & 6 & 228 & 0.94 & 0.75 & 0.24 & 0.76 & 0.80    \\
9 & 1533 & 97.86 & 5 & 137 & 0.88 & 0.50  & 0.13 & 0.87 & 0.69 \\
10 & 443  & 98.63 & 6 & 346 & 0.92 & 0.75 & 0.38 & 0.62 & 0.76   \\ 
......& .....&....& .....& .....& .....& .....& .....& ....& ....\\
100 & 384  & 100   & 5 & 296 &1	&0.50	&0.32 &	0.68 &	0.67 \\
\bottomrule
\end{tabular}
\end{table*}
We calculate the Aesthetic Score using a weighted average by evaluating Color Harmony, Lightness, and Simplicity (opposite to \textit{Complexity}, \textit{Simplicity = 1 - Complexity}). To calculate the measure, it is necessary to normalize the input data so that the range is between 0 and 1 (Eq. \ref{norm}). After that, the aesthetic score can be determined using Eq. \ref{wa}. 

\begin{equation}
    x_{\text{normalized}} = \frac{x - \min(x)}{\max(x) - \min(x)}
    \label{norm}
\end{equation}

\begin{equation}
    \text{Weighted Average} = \frac{\sum_{i=1}^{n} w_i \times x_i}{\sum_{i=1}^{n} w_i}
    \label{wa}
\end{equation}
where \(x_i\) represents the value of the \(i\)-th image feature value, \(w_i\) represents the weight associated with the \(i\)-th item, and \(n\) is the total number of items. We want to give more importance to lightness, so we assign it a double weight due to its critical importance in visual comfort \cite{light}. 

Table \ref{dataset_aes} shows the dataset updated with normalized values for \textit{CH, L, C} and \textit{Simplicity} along with the \textit{Aesthetics Score}. Notably, a weak but positive correlation (0,19) exists between the number of likes and the \textit{Aesthetic score} of interior design without accounting for individual color preferences.

\subsubsection{Single Color Preference}
\begin{figure}[tb]
\centering
\includegraphics[width=0.5\textwidth]{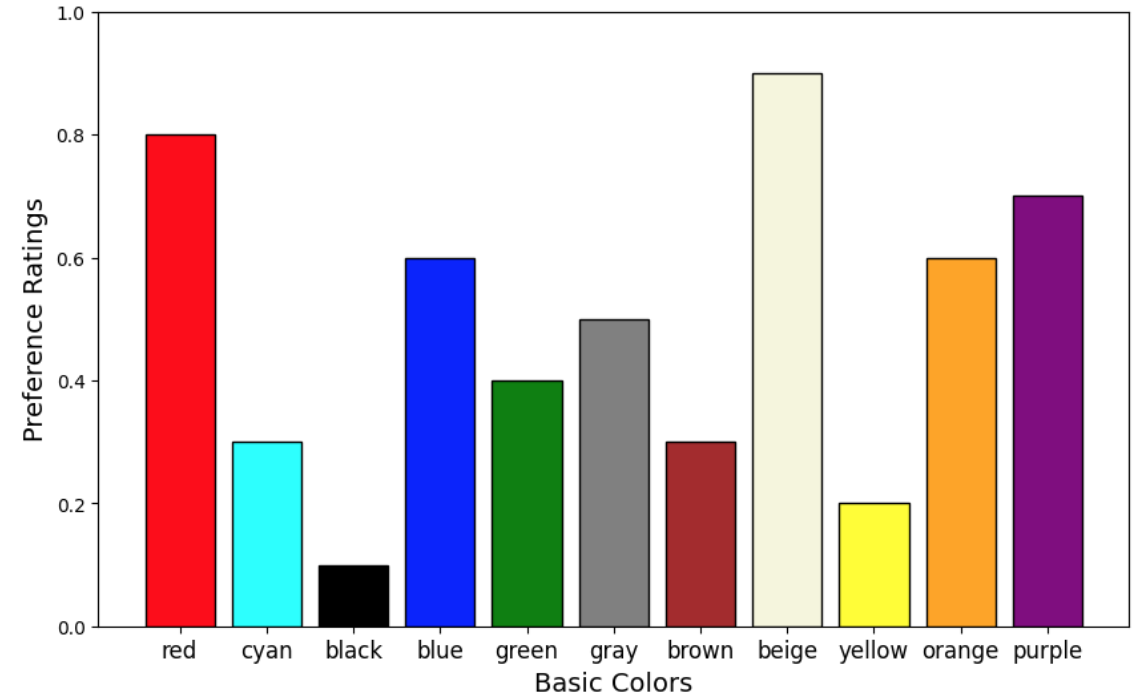}

\caption{Example of Single Color Ratings}
\label{fig_single_cr}
\end{figure}

Single Color Preference demonstrates the preference for single colors without context. People typically prefer certain colors over others, and almost everyone has a favorite color, such as green \cite{fss}. Fig. \ref{fig_single_cr} shows an example of single color ratings.
\begin{equation}
    Pref(i) = \frac{\sum_{i=1}^{5} Pref(Color_i) \times p_{Color_i}}{\sum_{i=1}^{5} p_{Color_i}}
    \label{bigeq}
\end{equation}

In Eq. \eqref{bigeq}, \( p_{Color_i} \) represents the pixel importance of the \( i \)-th color, and \( Pref(Color_i) \) are the user preferences for the individual colors. This equation calculates the weighted average of the preferences for the five colors.



\subsubsection{Fuzzy Inference System}
The proposed Fuzzy Inference System (FIS) aims to detect the user's aesthetic preference using two main fuzzy input variables: \textit{Aesthetic Score} and \textit{Color Preference}. The output fuzzy variable is \textit{Total preference}, reflecting the personalized aesthetic value. The inputs are partitioned into \textit{'Low', 'Medium'}, and \textit{'High'} linguistic terms. The output is defined by terms such as \textit{'Weak', 'Neutral', 'Strong'}, and \textit{'Very Strong'}, mirroring the inherent variability in preference levels. Fuzzy sets for input and output variables are shown in Fig. \ref{many_examples_palettes}. 

We define our model using nine specific fuzzy rules, as detailed in Table \ref{fuzzyrules}.  As a result, a fuzzy control system is created, followed by simulation. Input values are set, and the output is computed. For example, Rule 7: IF the \textit{Aesthetic Score} is \textit{Medium} AND \textit{Color Preference} is \textit{High} THEN \textit{Total Preference} is  \textit{Strong }for this interior design.

\begin{table}[tb]
\centering
\caption{Fuzzy rules used in the fuzzy inference system}
\label{fuzzyrules}
\begin{tabular}{cccc}
\toprule
\textbf{Rule} & \textbf{Aesthetic Score} & \textbf{Color Preference} & \textbf{Total Preference} \\
\midrule
1 & Low & Low & Weak \\
2 & Low & Medium & Weak \\
3 & Low & High & Neutral \\
4 & Medium & Low & Neutral \\
5 & Medium & Medium & Neutral \\
6 & High & Low & Neutral \\
7 & Medium & High & Strong \\
8 & High & Medium & Strong \\
9 & High & High & Very Strong \\
\bottomrule
\end{tabular}
\end{table}

\section{Results} 
\subsection{Examples}

Fig. \ref{fig_single_cr} displays an example of extracted basic colors from an interior design image. Methodology for basic color extraction was adapted from recent research \cite{2023coloremotion}.
\begin{figure}[!t]
\centering
\includegraphics[width=0.4\textwidth]{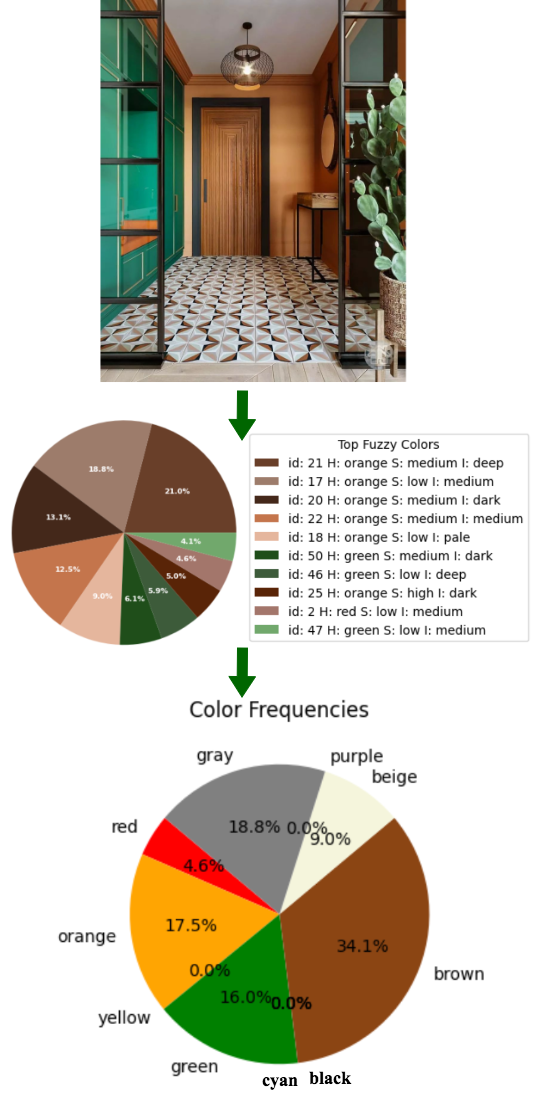}
 \caption{Extracting basic colors from image for color preference estimation}
\label{fig_single_cr}
\end{figure}

\begin{figure*}
  \begin{subfigure}{0.32\textwidth}
    \includegraphics[width=\linewidth]{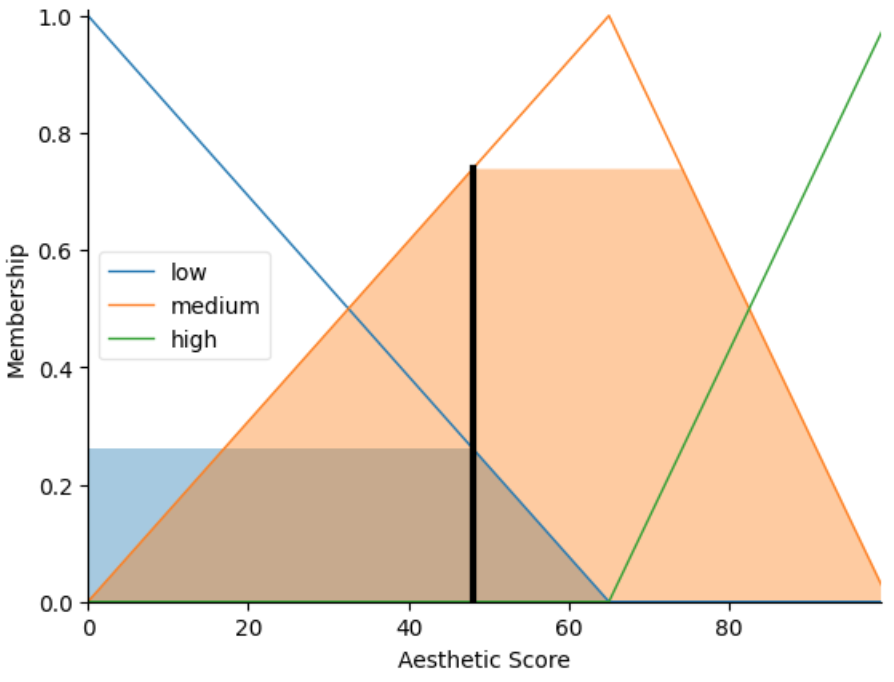}
    \caption{Applying input 48\% on \textit{Aesthetic Score} fuzzy set} \label{fig:1a}
  \end{subfigure}%
  \hspace*{\fill}   
  \begin{subfigure}{0.32\textwidth}
    \includegraphics[width=\linewidth]{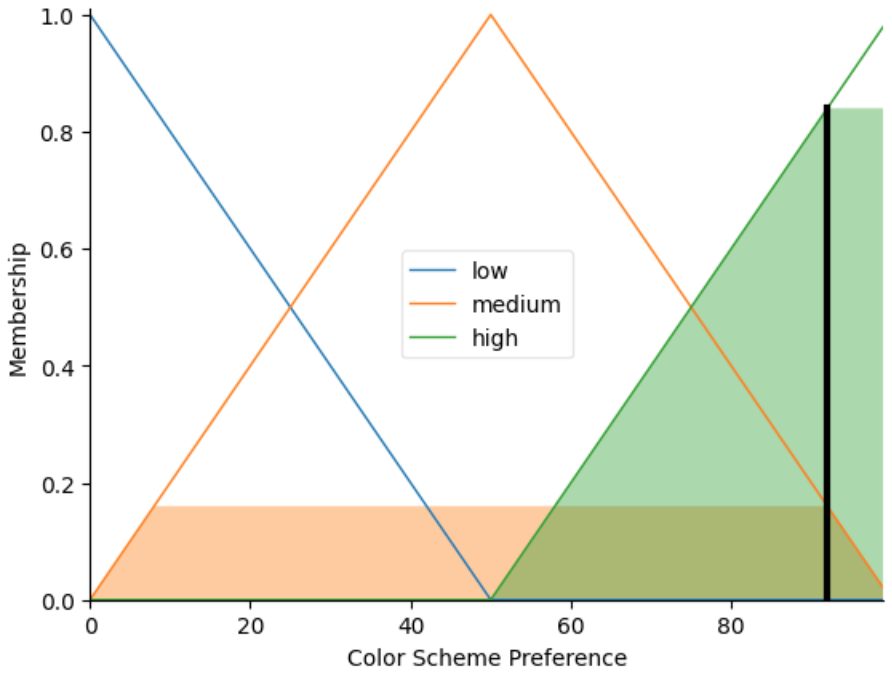}
    \caption{Applying input 92\% on \textit{Color Scheme Preference} fuzzy set} \label{fig:1b}
  \end{subfigure}%
  \hspace*{\fill}   
  \begin{subfigure}{0.32\textwidth}
    \includegraphics[width=\linewidth]{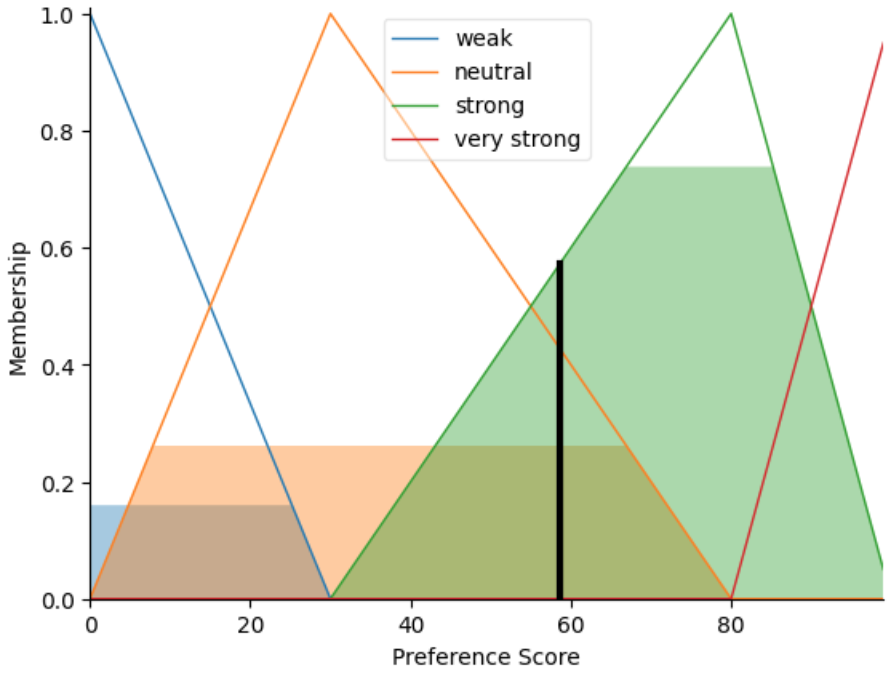}
    \caption{Aggregated Membership and Result, 58.72\%} 
  \end{subfigure}
\caption{Simulation Results.} \label{fig_ex}
\end{figure*}

\begin{figure*}
  \includegraphics[width=\textwidth]{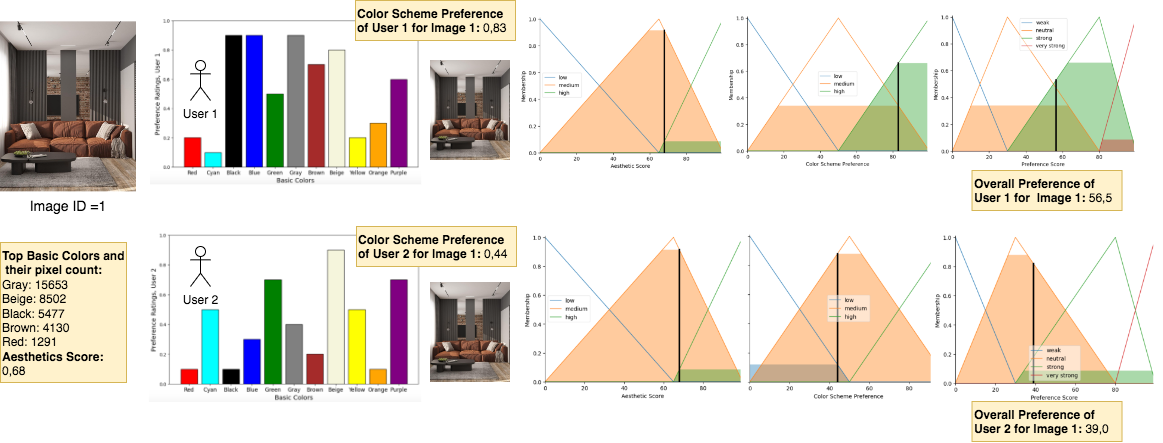}
\caption{Example of fuzzy inference. The Image 1 has an aesthetic score of 0.68, obtained by integrating \textit{CH, L, C}. The top five dominant colors are gray, beige, black, brown, and red. The respective color scheme preference for the selected design is 0.83 and 0.44. Finally, FIS computes the preferences of $User_{1}$ and $User_{2}$, which are 56.5 and 39.0, respectively.}
\label{figFuzzyres}
\end{figure*}

We must specify the inputs and apply the defuzzification method to simulate our fuzzy system. For illustration, let's consider the following situation: the \textit{Aesthetic Score} and \textit{Color Scheme Preference} are 48\% and 92\%, respectively (see Fig. \ref{fig_ex}. Subsequently, fuzzy aggregation combines the output membership functions through the maximum operator. Next, we must perform defuzzification to obtain a total preference; to do this, we employ the centroid method. The overall aesthetic preference is 58.72 \% as a result of aggregation based on fuzzy rules. The result, as visualized, is shown in Fig. \ref{fig_ex}.

Fig. \ref{figFuzzyres} shows an example of the inference process and how aesthetic scores are calculated. We can observe that $User_{1}$ and $User_{2}$ have vastly different individual color preferences. The sample interior design image has a general aesthetic score of 0.68, obtained by integrating \textit{CH, L, C}. The top five dominant colors fetched for this image are gray, beige, black, brown, and red. Considering their frequencies and single color ratings of $User_{1}$ and $User_{2}$, their respective color scheme preference for the selected design is 0.83 and 0.44. Finally, fuzzy inference systems use aesthetic scores of 0.68 and color preferences of 0.83 and 0.44 to compute the preferences of $User_{1}$ and $User_{2}$, which are 56.5 and 39.0, respectively. This notable difference in total aesthetic preference can be explained by the fact that almost all of the selected design's dominant colors are among the favorite colors of 
$User_{1}$, in contrast to $User_{2}$ color preferences.

\begin{figure}[htbp]
\centerline{\includegraphics{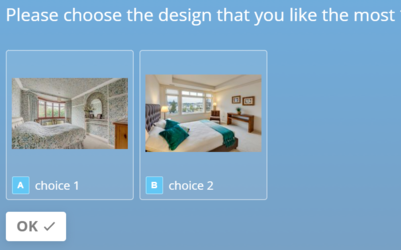}}
\caption{2AFC example}
\label{fig}
\end{figure}

The aesthetic evaluation results were presented using the fuzzy inference system for arbitrary interior design images.
Possible applications of our method can be:
\begin{itemize}
    \item Design Generation
    \item Design Enhancements
    \item Aesthetic Measurement of the design
    \item E-commerce Recommendation System (shopping coordinator that suggests specific items to add to an existing room to maximize aesthetic attractiveness).
\end{itemize}

\subsection{Performance Evaluation}

\begin{table}[tb]
\centering
\caption{Overall Preference of User (1,2) based on Aesthetic Score and Color Preference of each user}
\label{tab:your_table}
\begin{tabular}{@{}cccccc@{}}
\toprule
\multirow{2}{*}{\textbf{Image ID}} & \multirow{2}{*}{\textbf{Aesthetics Score}} & \multicolumn{2}{c}{\textbf{Color Preference}} & \multicolumn{2}{c}{\textbf{Overall Preference}}      \\ 
\cmidrule(lr){3-4} \cmidrule(lr){5-6}
                                   &                                            & \textbf{User1}    & \textbf{User2}    & \textbf{User1} & \textbf{User2}   \\ 
\midrule
1                                  & 0.68                                     & 0.83             & 0.44             & 56.5             & 39.0                 \\
91                                 & 0.52                                       & 0.64             & 0.37             & 43.9             & 36.5               \\
35                                 & 0.98                                     & 0.71             & 0.34             & 67.0               & 57.2               \\
86                                 & 0.74                                     & 0.84             & 0.46             & 57.6             & 43.8               \\
97                                 & 0.69                                     & 0.58             & 0.53             & 41.1             & 39.7               \\
\bottomrule
\end{tabular}
\end{table}

The method we used, called 2AFC (Two-Alternative Forced Choice), is a widely accepted approach in psychology and neuroscience \cite{2afc}. 2AFC method is a more efficient approach for identifying aesthetic preferences \cite{2Afcaesth}.  In a typical 2AFC trial, participants choose which of two displayed visuals they find more aesthetically pleasing. This process is repeated for all possible pairs of stimuli. So, 2AFC measures people's preferences by looking at their choices. The formula for the number of trials (T) needed in a 2AFC experiment to measure preferences for n stimuli is given by: \[ T = \frac{n \cdot (n - 1)}{2} \]
With 2 participants and five interior design images chosen (represented by \( n \)),  we get a total of 10 trials (\( T \)). In the context of the 2AFC method, the hit rate refers to the percentage or proportion of trials in which the participant correctly identifies the preferred or more aesthetically pleasing option among the two presented alternatives. 
In our case, the hit rate is calculated based on the number of correctly predicted interior design choices out of the total number of trials. The formula for the hit rate is as follows:
\begin{equation}
\text{Hit Rate} = \frac{\text{Number of Correctly Predicted} \\ \text{ Choices}}{\text{Total Number of Trials}}
\end{equation}
The percentage of correct predictions can be used to evaluate overall performance. The calculated hit rate is 0.7 (14 correct predictions out of 20). Experimental results are given in Table \ref{tab:your_table}, and the overall process is demonstrated in Fig. \ref{figFuzzyres}.

\section{Conclusion}


The current paper presents a novel approach to evaluating the personalized aesthetics score of an interior design considering color preferences and a general aesthetic score obtained based on visual features such as color harmony, lightness, and complexity.

This research supports previous studies that have endorsed the 2AFC task as the favored measurement technique for assessing aesthetic preferences \cite{2Afcaesth}, \cite{Palmer2013}. Our findings on the average hit rate of the 2AFC method in predicting aesthetic scores were similar to recent studies \cite{fss}, \cite{shamoi2023universal}.


The proposed approach has certain limitations, like the relatively small dataset for aesthetic feature extraction and the small number of participants in the 2AFC experiment. As for future works, this algorithm can be used for aesthetic assessment in other areas, such as fashion clothing compatibility, car aesthetics, website color classifier, etc. The practical significance is reflected in the model's potential applications in content creation, recommendation systems, and human-computer interaction, emphasizing its ability to align with human aesthetic perceptions as revealed through 2AFC. 


This research can assist designers and experts in better understanding and meeting people's interior design preferences, particularly in today's digitally driven environment.



\bibliography{export}

\end{document}